\PassOptionsToPackage{a4paper,margin=1in}{geometry}
\documentclass[pdflatex,sn-mathphys-num]{sn-jnl}

\usepackage{graphicx}%
\usepackage{multirow}%
\usepackage{amsmath,amssymb,amsfonts}%
\usepackage{amsthm}%
\usepackage{mathrsfs}%
\usepackage[title]{appendix}%
\usepackage{xcolor}%
\usepackage{textcomp}%
\usepackage{manyfoot}%
\usepackage{booktabs}%
\usepackage{algorithm}%
\usepackage{algorithmicx}%
\usepackage{algpseudocode}%
\usepackage{listings}%
\usepackage{longtable}
\usepackage{tabularx}
\usepackage{lscape}
\usepackage{booktabs}
\usepackage{adjustbox}
\usepackage{array}
\usepackage{caption}

\theoremstyle{thmstyleone}%
\theoremstyle{thmstyletwo}%

\theoremstyle{thmstylethree}%

\begin{document}

\title[Article Title]{Integrating Reinforcement Learning with Visual Generative Models: Foundations and Advances}


\author[1]{\fnm{Yuanzhi} \sur{Liang}}\email{liangyzh18@outlook.com}

\author[1]{\fnm{Yijie} \sur{Fang}}\email{fangyijie@stu.xidian.edu.cn}\equalcont{These authors contributed equally to this work and are listed in alphabetical order by last name.}

\author[1]{\fnm{Ke} \sur{Hao}}\email{ke\_hao2002@outlook.com}\equalcont{These authors contributed equally to this work and are listed in alphabetical order by last name.}

\author[1]{\fnm{Rui} \sur{Li}}\email{rui.li@mail.ustc.edu.cn}\equalcont{These authors contributed equally to this work and are listed in alphabetical order by last name.}

\author[1]{\fnm{Ziqi} \sur{Ni}}\email{zqni@seu.edu.cn} \equalcont{These authors contributed equally to this work and are listed in alphabetical order by last name.}

\author[1]{\fnm{Ruijie} \sur{Su}}\email{472171770@qq.com} \equalcont{These authors contributed equally to this work and are listed in alphabetical order by last name.}

\author[1]{\fnm{Chi} \sur{Zhang}}\email{zhangc120@chinatelecom.cn}

\affil[1]{Institute of Artificial Intelligence (TeleAI), China Telecom}

\abstract{Generative models have made significant progress in synthesizing visual content, including images, videos, and 3D/4D structures. However, they are typically trained with surrogate objectives such as likelihood or reconstruction loss, which often misalign with perceptual quality, semantic accuracy, or physical realism.
Reinforcement learning (RL) offers a principled framework for optimizing non-differentiable, preference-driven, and temporally structured objectives. Recent advances demonstrate its effectiveness in enhancing controllability, consistency, and human alignment across generative tasks.
This survey provides a systematic overview of RL-based methods for visual content generation. We review the evolution of RL from classical control to its role as a general-purpose optimization tool, and examine its integration into image, video, and 3D/4D generation. Across these domains, RL serves not only as a fine-tuning mechanism but also as a structural component for aligning generation with complex, high-level goals. We conclude with open challenges and future research directions at the intersection of RL and generative modeling.
}





\maketitle

\footnotetext{Survey homepage: \url{https://visgenrlsurvey.liangyzh18.workers.dev/}}

\section{Introduction}\label{sec_intro}

Recent advances in generative modeling—particularly diffusion models~\cite{yang2023diffusion,croitoru2023diffusion} and autoregressive approaches~\cite{xiong2024autoregressive,kaur2023autoregressive}—have substantially enhanced the quality of synthesized content across image~\cite{dhariwal2021diffusion,rombach2022high}, video~\cite{rdpo,seedance}, and 3D contents~\cite{ye2024dreamreward,lin2020modeling}. These breakthroughs have increasingly demonstrated the potential to extend beyond content generation into broader AI paradigms~\cite{liang2019vrr, AIFlowReport, chen2024hydra, PN, VPN, PiNDA, PiNI, PiNGDA, MiN}. 
Despite this progress, most generative models are trained using surrogate objectives such as maximum likelihood estimation or reconstruction loss, which are often misaligned with human perception~\cite{zhang2018unreasonable, esser2021taming}, semantics~\cite{radford2021learning}, or physical plausibility~\cite{ho2022video}. As a result, generated outputs can exhibit artifacts such as motion inconsistency~\cite{villegas2017learning, yan2021videogpt}, or structural errors~\cite{mescheder2018training}.


Reinforcement learning (RL)~\cite{sutton1998reinforcement} offers a principled framework for optimizing non-differentiable, human-aligned, or environment-driven objectives through interaction. Originally designed for control and decision-making, RL has emerged as a powerful complement to generative models, enabling long-horizon optimization, feedback-aware learning, and preference-based fine-tuning. In recent works~\cite{dhariwal2021diffusion,dancegrpo,zhou2024flowgrpo}, RL has been increasingly adopted to improve controllability, enhance realism, and align generation with complex and structured objectives across multiple modalities.

\begin{figure}[t!]
    \centering
    \includegraphics[width=0.6\textwidth]{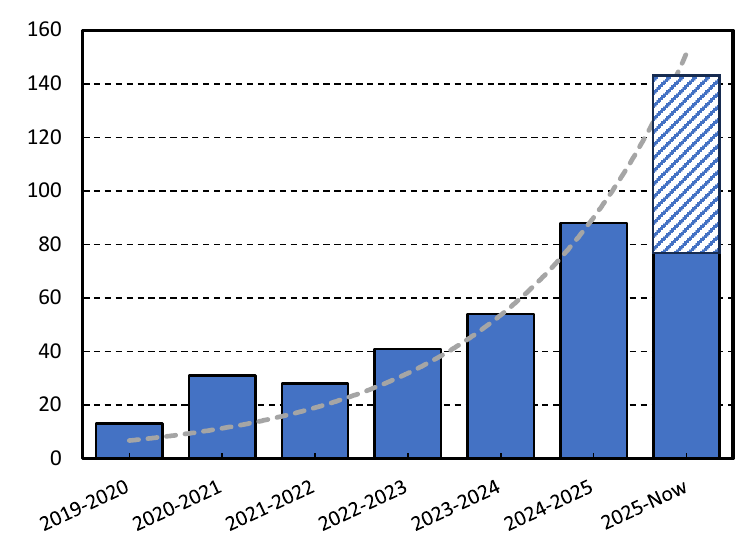}
    \caption{Growth of publications at the intersection of reinforcement learning and visual content generation (2019–2025). The field has experienced exponential growth, increasing from 13 papers in 2019–2020 to 91 in 2024–2025 (as of July 30). With 77 papers already published in the first half of 2025, the year is projected to exceed 140 publications. This trend reflects the field’s transition from exploration to consolidation and its growing strategic importance in visual generation research.}
    \label{fig:paper_sta}
\end{figure}

The convergence of RL and generative modeling is experiencing rapid growth. As illustrated in Figure~\ref{fig:paper_sta}, the number of research papers at this intersection has risen sharply—from just 13 publications in 2019–2020 to 91 in 2024–2025 (as of July 30). With 77 papers already published in the first half of 2025, this year is projected to exceed 140 publications. This exponential rise reflects a transition from early exploration to a more systematic and widely recognized research paradigm, indicating the growing strategic importance of RL-enhanced generation in the broader AI landscape.

This survey presents a comprehensive overview of how RL is being integrated into modern generative modeling across multiple domains. We begin with the conceptual evolution of RL—from its origins in solving Markov decision processes to its emergence as a general-purpose optimization framework embedded within broader learning systems. We then examine its application in image generation, where RL techniques are used to improve semantic alignment, aesthetic quality, and controllability through policy optimization and preference-based learning. In the video domain, RL contributes to efficient sampling, temporal coherence, physical plausibility, and alignment with user intent. Finally, we explore the use of RL in 3D content generation, including point cloud reconstruction, mesh synthesis, multi-view consistency, and human motion modeling. In all, these threads illustrate the growing role of RL as a versatile and principled mechanism for guiding generative models toward complex, structured, and human-aligned objectives.


Across these domains, RL serves as a flexible and interpretable optimization layer that complements existing architectures. By reframing generation as an interactive process driven by evaluative feedback, RL allows generative models to move beyond static supervision and toward learning from preferences, outcomes, or real-world constraints. This survey aims to map the methodological landscape, identify cross-domain patterns, and offer insights into future research directions at the intersection of RL and generative modeling.

\section{RL Revolution}\label{sec_rl}

Reinforcement learning (RL)~\cite{sutton1998reinforcement} provides a general framework for sequential decision-making, where an agent learns to act by interacting with an environment to maximize long-term rewards. It is uniquely characterized by delayed feedback and the need to trade off exploration and exploitation. In the past decade, RL has demonstrated remarkable success in simulated domains such as video games~\cite{mnih2015human,shao2019survey}, robotics~\cite{ibarz2021train,zhao2020sim,kalashnikov2018scalable}, and continuous control~\cite{dulac2021challenges,lillicrap2015continuous}, showcasing its potential under well-defined dynamics and reward structures~\cite{tassa2018deepmind,mnih2015human,wang2024distributionally}. Yet, real-world applications remain limited. Practical deployments often suffer from sample inefficiency~\cite{mohanty2021measuring,kaiser2019model}, reward specification challenges~\cite{ng1999policy,christiano2017deep}, and poor generalization~\cite{zhang2018study,arjovsky2020out}. The diversity of real-world environments~\cite{dulac2019challenges,zhang2018study} further complicates the development of universally effective algorithms.

To better understand RL’s evolving role, this survey adopts a broader perspective: rather than viewing RL as a fixed set of algorithms, we frame it as a general paradigm for learning from interaction, feedback, and preferences. We organize this evolution into four phases, culminating in its emerging integration with generative modeling and human-aligned learning.

Phase I: RL as solving Markov Decision Processes (MDPs)~\cite{bellman2015applied,puterman2014markov} through trial-and-error~\cite{sutton1991dyna,sutton1998reinforcement}.

Phase II: A fragmentation into subfields (e.g., offline RL~\cite{kumar2020conservative,fujimoto2019off}, model-based RL~\cite{chua2018deep,hafner2019dream}) to overcome specific bottlenecks.

Phase III: A redefinition of RL’s core problem—shifting from policy optimization in known environments to learning environment dynamics~\cite{schrittwieser2020mastering} and aligning with human intent~\cite{christiano2017deep,ouyang2022training}.

Phase IV: The emergence of RL as a general substrate for decision-making~\cite{hafner2023mastering}, interfacing with planning~\cite{silver2018general}, simulation~\cite{ha2018worldmodels}, and generative modeling~\cite{dancegrpo,rdpo}.

\subsection{RL as MDP Solvers}
The first phase of RL was grounded in the formalism of Markov Decision Processes (MDPs), which define sequential decision-making under uncertainty through a set of states, actions, transition dynamics, and reward functions~\cite{bellman2015applied,puterman2014markov,bellman1957dp}. The goal was to compute an optimal policy that maximizes expected cumulative reward. Early RL efforts closely mirrored the dynamic programming solutions for MDPs, assuming either full or sampled access to the environment.

Two major solution paradigms emerged: value-based and policy-based methods. Value-based approaches estimate the expected return (value) of actions or states and derive policies by acting greedily with respect to these estimates. In contrast, policy-based methods directly parameterize and optimize the policy itself, often through gradient ascent on expected reward.

Value-based methods were historically dominant. Temporal-Difference (TD) learning~\cite{sutton1988td} laid the foundation by updating value estimates from sampled transitions, blending Monte Carlo evaluation with dynamic programming. Algorithms such as Q-learning~\cite{watkins1992q} and SARSA~\cite{rummery1994line} introduced off-policy and on-policy learning variants, while tabular implementations were standard for small-scale problems.

Policy-based methods gained traction for their natural handling of stochastic policies and continuous action spaces. The REINFORCE algorithm~\cite{williams1992simple} introduced Monte Carlo policy gradients using the likelihood ratio trick, albeit with high variance. This led to actor-critic architectures, which paired a learned value function (critic) to reduce the variance of policy updates.

The integration of deep learning dramatically extended the reach of both paradigms. The Deep Q-Network (DQN)~\cite{mnih2015human} approximated the Q-function with a convolutional neural network and achieved human-level performance on Atari games using raw pixel inputs. Key enhancements such as Double DQN~\cite{hasselt2016deep}, Dueling networks~\cite{wang2016dueling}, and prioritized replay~\cite{schaul2016prioritized} were later unified in the Rainbow agent~\cite{hessel2018rainbow}. In parallel, scalable policy-based methods emerged. Asynchronous Advantage Actor-Critic (A3C)\cite{mnih2016asynchronous} introduced parallelism and advantage estimation; Trust Region Policy Optimization (TRPO)\cite{schulman2015trust} and its practical variant Proximal Policy Optimization (PPO)~\cite{schulman2017proximal} stabilized updates with theoretically grounded constraints, becoming standard in continuous control.

These advances solidified deep RL as a powerful tool for solving complex MDPs, with notable successes in game playing~\cite{silver2016mastering}, robotic manipulation~\cite{levine2016end}, and simulated control~\cite{tassa2018deepmind}. However, the MDP-centric paradigm revealed several fundamental limitations that shaped the next phase of research. (1) Sample inefficiency. Most algorithms required millions of environment interactions to learn effective policies—tractable in simulation but prohibitive in domains like robotics or healthcare. This inefficiency arose from poor data reuse, slow policy updates, and especially inefficient exploration, which struggled in sparse-reward scenarios where agents failed to discover meaningful trajectories or assign credit over long horizons. (2) Limited generalization. RL agents were typically evaluated in the same environments they were trained on, leading to overfitting to visual or structural artifacts. Benchmarks like Procgen revealed that even small perturbations could significantly degrade performance, highlighting poor robustness and weak transferability. (3) Unrealistic assumptions. Classical RL methods often relied on fully observable, stationary environments with well-specified rewards and unrestricted interaction—conditions rarely met in practice. In many real-world tasks, interaction is costly or unsafe, data is offline or limited, and objectives are implicit or multi-faceted, making these assumptions untenable.

These limitations prompted a shift in focus from solving fixed MDPs to relaxing their assumptions—leading to the rise of offline RL, model-based methods, and alignment-driven learning. This transition marks the beginning of Phase II.

\subsection{Emergence of Specialized Subfields} 
Offline RL, model-based RL, and other subfields (multi-agent RL, safe RL, etc.) emerged as semi-independent research areas. Each introduced new assumptions or additional components to the RL paradigm to tackle specific challenges: removing the need for active environment interaction, incorporating knowledge of dynamics, handling multiple agents or risk, and so on. This fragmentation represented a paradigm shift: rather than viewing RL as just an agent solving an MDP, RL could be re-formulated to fit constraints of data availability, environment knowledge, or task structure.

\noindent \textbf{Offline RL:}
Offline RL addresses the challenge of learning effective policies from fixed datasets without any further environment interaction. This setting is particularly valuable in domains where exploration is costly, risky, or infeasible—such as autonomous driving, healthcare, and industrial control. By reframing RL as a static-data learning problem, offline RL brings it closer to supervised learning, enabling more scalable and safer deployment in real-world systems.

However, this paradigm introduces a central challenge: distributional shift between the behavior policy that collected the data and the target policy being learned. Without access to the environment, agents may assign erroneously high value to out-of-distribution actions, leading to extrapolation error and unstable training~\cite{kumar2020conservative}. Recent methods address this through conservative value estimation or policy regularization. For example, CQL~\cite{kumar2020conservative} penalizes Q-values for unseen actions to avoid overestimation, while BRAC~\cite{wu2019behavior} and TD3+BC~\cite{fujimoto2021minimalist} constrain the learned policy to remain close to the data distribution. IQL~\cite{kostrikov2021offline} avoids explicit constraints but modifies the Bellman update to implicitly favor in-distribution actions. These advances, along with growing theoretical understanding of bootstrapping error and pessimism, have made offline RL a promising solution for safe, sample-efficient learning from static logs—while also raising new challenges in data coverage, evaluation, and generalization.

\noindent \textbf{Multi-Agent RL:} When multiple agents interact in the same environment, the single-agent MDP framework no longer holds because the environment becomes non-stationary from each agent’s perspective. This challenge led to the development of multi-agent RL (MARL)~\cite{canese2021multi,zhang2021multi,bucsoniu2010multi,foerster2018counterfactual}, which addresses issues such as non-stationarity, cooperation, competition, and equilibrium strategies. Key contributions include MADDPG~\cite{lowe2017multi}, which introduced centralized training with decentralized execution, allowing agents to condition their critics on global information during training while executing independently at test time. Other approaches, such as LOLA~\cite{foerster2017learning}, incorporated opponent modeling by anticipating how an agent's updates influence others' learning. MARL expanded the classical RL framework to stochastic games and multi-agent MDPs, drawing on concepts from game theory to develop stable learning dynamics and policy solutions.

\noindent \textbf{Risk-Sensitive and Robust RL:} Traditional RL optimizes the expected return, which may overlook performance variability or safety-critical failures. Risk-sensitive methods~\cite{chow2015risk} address this by optimizing alternative criteria, such as minimizing variance or ensuring favorable worst-case outcomes. Distributional RL models~\cite{bellemare2017distributional,dabney2018distributional} the entire return distribution rather than its mean, enabling richer decision-making. Robust RL~\cite{pinto2017robust,vinitsky2020robust,kamalaruban2020robust} further addresses uncertainty in dynamics or rewards, seeking policies that perform well under perturbations or adversarial changes. Techniques such as domain randomization~\cite{tobin2017domain} and adversarial training~\cite{kamalaruban2020robust} have been used to enhance robustness and stress-test policies. These approaches reflect a shift from reward maximization toward more reliable and resilient objectives.

\noindent \textbf{Safe RL:} Safe RL focuses on satisfying explicit safety constraints throughout training and deployment. It often formulates the problem as a constrained MDP, incorporating methods such as Lagrangian optimization~\cite{achiam2017constrained}, reward shaping~\cite{garcia2015comprehensive}, or safety shields~\cite{alshiekh2018safe,elsayed2021safe} to enforce limits on unsafe behavior. This is particularly critical in domains like autonomous driving~\cite{zhao2024survey} and robotics~\cite{berkenkamp2017safe}, where unsafe exploration can lead to catastrophic failures. Safe RL also intersects with exploration research, aiming to develop learning strategies that remain within safety boundaries while still enabling policy improvement.

\subsection{Redefining the Problem — RL as Learning to Simulate, Optimize, and Align} 
During the late 2010s to mid-2020s, RL evolved from optimizing fixed reward functions in known environments to serving as a modular component within broader learning systems. This phase emphasized two key directions: first, using RL to optimize over models learned from data, such as simulators or reward functions~\cite{schrittwieser2020mastering,hafner2019dream}; second, aligning agent behavior with nuanced human goals that cannot be easily captured by scalar rewards~\cite{christiano2017deep,ouyang2022training}. The focus shifted from refining policy optimization algorithms to designing the outer structure of the learning task itself, including how preferences are modeled and how environment dynamics are inferred~\cite{ziegler2019fine,bai2022training}. RL became less about the optimizer and more about how the problem is formulated.

\noindent \textbf{RL from Human Feedback: }
RL from Human Feedback (RLHF)~\cite{christiano2017deep,ouyang2022training} has become a high-impact application of RL, particularly in aligning large language models with human intentions. Rather than relying on hand-crafted reward functions, RLHF uses human preferences to define the objective. The typical pipeline involves collecting human demonstrations or rankings of model outputs~\cite{christiano2017deep,ziegler2019fine}, training a reward model to predict these preferences, and then fine-tuning the model using RL to maximize this learned reward~\cite{ouyang2022training}. This transforms vague goals like “be helpful” into a concrete RL problem: the model generates responses, the reward is given by the preference model, and RL optimizes accordingly.

The success of InstructGPT~\cite{ouyang2022training} demonstrated the effectiveness of this approach. A 1.3B parameter model fine-tuned via RLHF outperformed the original 175B GPT-3 in instruction following, while reducing harmful or irrelevant outputs. RLHF builds on earlier work such as preference-based RL~\cite{christiano2017deep}, which showed that reward functions can be learned from human comparisons rather than manually specified. In high-dimensional spaces like language or vision, such feedback provides a scalable supervision signal. Importantly, the core challenge in RLHF lies not in the RL algorithm itself, but in designing robust reward models from subjective data~\cite{bai2022training}. This paradigm has since extended to domains like image generation and robotics, where human-aligned behavior is critical~\cite{ziegler2019fine}. Overall, RLHF exemplifies a broader view of RL—as a flexible framework for aligning complex models with implicit goals through interaction and evaluative feedback.

\noindent \textbf{World Models and the Resurgence of Model-Based RL:} 
Model-based RL has re-emerged as a promising solution to the sample inefficiency of traditional RL~\cite{ha2018worldmodels, chua2018deep}. By learning a model of environment dynamics from data, agents can simulate interactions internally, enabling policy training with significantly fewer real-world interactions. Early work such as World Models and PETS demonstrated that policies trained in learned simulators could transfer successfully to real environments~\cite{ha2018worldmodels, chua2018deep}, highlighting the potential of model-based approaches.

This direction was further advanced by Dreamer and its successors, which learned latent dynamics models and trained agents entirely within imagined trajectories~\cite{hafner2019dream, hafner2020dreamerv2}. DreamerV2~\cite{hafner2020dreamerv2} notably achieved competitive performance on the Atari benchmark using far fewer environment frames, demonstrating scalability to high-dimensional, vision-based tasks~\cite{hafner2020dreamerv2}. MuZero took a different approach by learning an implicit model to support planning via tree search, blending learning and planning without requiring a full transition model~\cite{schrittwieser2020mastering}.

Recent architectures such as DreamerV3 and Diffusion-QL integrate world models with generative and preference-aware objectives, illustrating a shift in focus from algorithmic innovation to task specification and system-level design~\cite{hafner2023dreamerv3, wang2022diffusion}. In this view, RL increasingly involves first learning the underlying Markov decision process—its dynamics and rewards—followed by planning or policy optimization within that learned model. This two-stage formulation offers greater flexibility and data efficiency, and has proven effective even when the learned model is imperfect. The result is a broader, more adaptable vision of RL, where modeling, simulation, and optimization are deeply intertwined.

\subsection{RL as a General Substrate for Decision-Making}

Recently, RL is increasingly viewed not as a standalone technique but as a general-purpose framework for decision-making across diverse domains. Rather than focusing solely on interactive agents in fixed environments, modern RL is embedded into larger systems involving planning, simulation, supervised learning, and human feedback.

A key feature of this phase is the emphasis on problem formulation over algorithmic novelty. In applications like RLHF~\cite{ouyang2022training}, the choice of RL algorithm (e.g., PPO~\cite{engstrom2019implementation}, GRPO~\cite{shao2024deepseekmath,chen2025seed}) is secondary; the critical component is the learned reward model derived from human preferences. Similarly, in systems like AlphaGo~\cite{granter2017alphago}, RL serves as one module within a broader architecture that includes imitation, self-play, and tree search. In these cases, RL acts as a fine-tuning layer that optimizes a well-defined objective.

This shift reflects a broader trend: RL is now treated as a general decision optimization tool. It enables learning from implicit or non-differentiable objectives, facilitates alignment with complex preferences, and extends to structured prediction and generative modeling. Tasks traditionally tackled with supervised learning can often be reframed in RL terms, with rewards representing task-specific metrics.

As RL integrates more deeply with other learning paradigms, its role becomes both more abstract and more central. It is increasingly used to tune system behavior, guide architectural search, and even coordinate self-improvement. Rather than viewing RL as a method confined to interactive environments, Phase IV positions it as a unifying substrate—capable of driving goal-directed behavior in systems that learn, adapt, and align across modalities and objectives.

\begin{figure}
    \centering
    \includegraphics[width=1\linewidth]{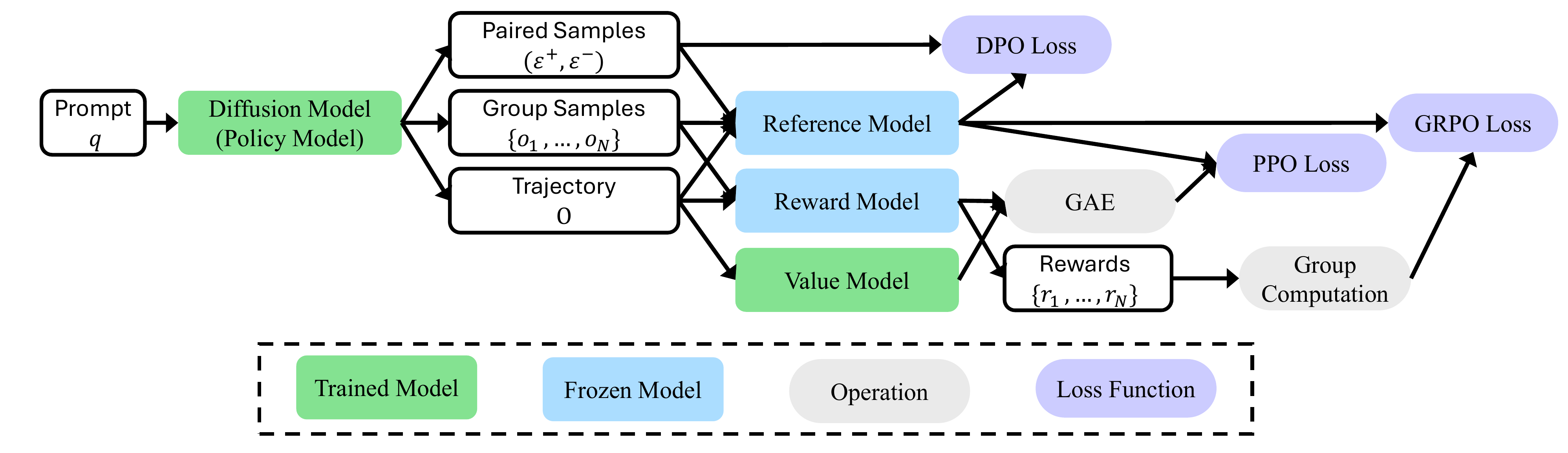}
    \caption{RL-style post-training pipelines for diffusion models.  Given a conditioning input $q$, the diffusion policy $\pi_\theta$ generates samples via iterative denoising.
DPO optimizes pairwise preferences using denoiser outputs (e.g., $\epsilon$) relative to a frozen reference policy.
PPO treats the denoising path $o$ as a policy rollout and updates the policy using advantage estimates. GRPO extends this formulation by computing relative advantages from a group of diffusion rollouts.
All methods employ KL regularization to the reference policy to stabilize training.}
    \label{fig:methodsummary}
\end{figure}

\section{RL for Image Generation}\label{sec_img}
Reinforcement learning (RL) offers a principled framework to steer generative models—particularly diffusion models—toward objectives that are difficult to optimize with standard supervised losses. These include non-differentiable metrics such as aesthetic quality, precise semantic alignment, and compliance with safety guidelines. RL addresses these challenges by reframing the generative process as a sequential decision-making task, where reward signals guide the model to satisfy complex, often preference-driven goals. As a result, RL provides an effective mechanism for guiding image generation models toward desired outcomes by optimizing such reward signals. Unlike purely supervised methods, RL excels at incorporating human preferences and maintaining semantic consistency between inputs (e.g., prompts or captions) and generated images. Current RL-based approaches for image generation fall into two main categories. The first includes policy-based methods, such as Proximal Policy Optimization (PPO) and recent variants like Group Relative Policy Optimization (GRPO), which optimize policies through gradient-based updates. The second is Direct Preference Optimization (DPO), which reformulates the problem as a preference classification task using ranked output pairs. A chronological overview of representative RL methods for image generation is provided in Table~\ref{tab:imgrl_methods_overview}.

\begin{table}[h!]
\centering
\caption{Comprehensive overview of RL methods applied to image generation, organized chronologically.}
\label{tab:imgrl_methods_overview}
\begin{tabular}{
    >{\raggedright\arraybackslash}p{3.4cm}
    >{\raggedright\arraybackslash}p{2.8cm}
    >{\raggedright\arraybackslash}p{2.5cm}
    >{\raggedright\arraybackslash}p{2.8cm}
}
\toprule
\textbf{Method} & \textbf{RL Approach} & \textbf{Publication Date} & \textbf{Venue} \\
\midrule
DPOK\cite{fan2023dpok}                   & Policy Gradient & May 2023  & NeurIPS 2023 \\
Promptist\cite{hao2023optimizing}        & Policy Gradient & Dec 2022  & NeurIPS 2023 \\
DDPO\cite{black2023training}             & Policy Gradient & May 2023  & Arxiv \\
RLD\cite{miao2024training}               & Policy Gradient & May 2023  & CVPR 2024 \\
Diffusion-DPO\cite{wallace2024diffusion} & DPO             & Nov 2023  & CVPR 2024 \\
PRDP\cite{deng2024prdp}                  & DPO             & Feb 2024  & CVPR 2024 \\
DenseRewardT2I\cite{yang2024dense}   & DPO             & Feb 2024  & ICML 2024 \\
POSI\cite{wu2024universal}               & Policy Gradient & Feb 2024  & Arxiv \\
AGFSync\cite{an2025agfsync}              & DPO             & Mar 2024  & Arxiv \\
PAE\cite{mo2024dynamic}                  & Policy Gradient & Apr 2024  & CVPR 2024 \\
CurriculumDPO\cite{croitoru2025curriculum} & DPO           & May 2024  & CVPR 2025 \\
HG-DPO\cite{na2025boost}                 & DPO             & May 2024  & CVPR 2025 \\
SPO\cite{liang2024step}                  & DPO             & Sat 2024  & Arxiv \\
ReNO\cite{eyring2024reno}                & DPO             & Jun 2024  & NeurIPS 2024 \\
DUO\cite{park2024direct}                 & DPO             & Jul 2024  & NeurIPS 2024 \\
RankDPO\cite{karthik2024scalable}        & DPO             & Oct 2024  & Arxiv \\
PatchDPO\cite{huang2025patchdpo}         & DPO             & Dec 2024  & CVPR 2025 \\
PPD\cite{dang2025personalized}           & DPO             & Jan 2025  & CVPR 2025 \\
ImageGenerationCoT\cite{guo2025can}    & DPO             & Jan 2025  & Arxiv \\
CaPO\cite{lee2025calibrated}             & DPO             & Feb 2025  & CVPR 2025 \\
DesignDiffusion\cite{wang2025designdiffusion} & DPO        & Mar 2025  & CVPR 2025 \\
LightGen\cite{wu2025lightgen}            & DPO             & Mar 2025  & Arxiv \\
SimpleAR\cite{wang2025simplear}          & GRPO            & Apr 2025  & Arxiv \\
DanceGRPO\cite{dancegrpo}         & GRPO            & May 2025  & Arxiv \\
Flow-GRPO\cite{liu2025flow}              & GRPO            & May 2025  & Arxiv \\
T2I-R1\cite{jiang2025t2i}                & GRPO            & May 2025  & Arxiv \\
ReasonGen-R1\cite{zhang2025reasongen}    & GRPO            & May 2025  & Arxiv \\
\bottomrule
\end{tabular}
\end{table}

\subsection{PPO-based Image Generation}
PPO-based image generation represents an early exploration of RL for improving image synthesis, particularly in aligning model outputs with human preferences, enhancing semantic consistency, and enabling controllable generation. These approaches treat image generation, especially in diffusion or autoregressive model, as a sequential decision-making process and leverage learned reward signals to guide policy optimization.

One representative approach is Denoising Diffusion Policy Optimization (DDPO)~\cite{black2023training}, which formulates the denoising process in diffusion models as a multi-step MDP. This allows the model to directly optimize downstream generation objectives, rather than relying solely on likelihood-based training. Concurrently, DPOK~\cite{fan2023dpok} also advances this direction by framing the fine-tuning of text-to-image diffusion models as an online RL problem within a multi-step Markov Decision Process (MDP) framework. It optimizes feedback-trained rewards using policy gradient, critically incorporating Kullback–Leibler (KL) regularization to concurrently enhance both image quality and text-image alignment.
To address the sparse-reward challenge in such policy-gradient approaches, B²-DiffuRL \cite{hu2025towards} introduces a backward progressive training scheme with branch-based sampling, which densifies the learning signal and enables more precise policy updates, leading to better alignment-diversity trade-offs. 
Separately, LOOP \cite{gupta2025simple} enhances training stability and efficiency by employing a leave-one-out baseline across multiple trajectories, effectively reducing variance and memory overhead while preserving the clipping mechanism of PPO.


To promote output diversity, Wallace et al.~\cite{miao2024training} introduce reward functions based on Maximum Mean Discrepancy (MMD) and mutual information. These rewards evaluate the distributional coverage of generated images relative to reference sets. Each image receives marginal-utility-based feedback, encouraging the model to produce diverse outputs while maintaining sample quality.

Beyond generation-stage optimization, PPO-based methods have also been applied to prompt engineering. Promptist~\cite{hao2023optimizing} proposes a two-stage pipeline that combines supervised fine-tuning (SFT) with PPO-based RL to automatically optimize user prompts. This improves alignment between user intent and the model’s internal preference signals. POSI~\cite{wu2024universal} extends this idea to safe text-to-image generation by using a custom reward function that balances semantic alignment with safety constraints. PAE~\cite{mo2024dynamic} further introduces dynamic prompt optimization, enabling fine-grained control over prompt weights and injection timing. By integrating SFT with online PPO, PAE optimizes for aesthetics, semantic fidelity, and user preference under a unified reward formulation.

\subsection{DPO-based Image Generation}
Direct Preference Optimization (DPO) has emerged as a highly effective method for aligning image generation models with human preferences. Early work applied DPO to standard generative architectures. For example, Diffusion-DPO~\cite{wallace2024diffusion} and PRDP~\cite{deng2024prdp} align diffusion models with preference data through post-training optimization, improving both output quality and controllability.

DPO has also been extended to specific sub-tasks. HG-DPO~\cite{na2025boost} addresses structural and pose inaccuracies in human image generation by using high-quality real images as reference preferences. RPO introduces a harmonic reward and early stopping to improve theme-guided generation while reducing overfitting. PatchDPO~\cite{huang2025patchdpo} incorporates patch-level feedback for personalized generation, enhancing local consistency and detail fidelity. Similarly, DesignDiffusion~\cite{wang2025designdiffusion} applies DPO to produce high-quality design-oriented images from text prompts.

In addition to task-specific refinements, several works apply DPO to broader challenges in text-to-image generation. ReNO~\cite{eyring2024reno} improves compositional detail by optimizing initial noise during inference, helping models generalize beyond fine-tuning. DUO~\cite{park2024direct} reframes content safety as a preference optimization task, aiming to suppress unsafe outputs. SPO~\cite{liang2024step} focuses on aesthetic alignment by optimizing for visual appeal based on preference signals.

Some works refocus on enhancing the DPO framework itself. AGFSync~\cite{an2025agfsync} reduces dependence on manual labels by generating preference annotations via AI models. RankDPO~\cite{karthik2024scalable} and CaPO~\cite{lee2025calibrated} propose scalable modeling techniques to address the limitations of aging or costly human preference data. Dense Reward~\cite{yang2024dense} introduces dense reward signals to capture fine-grained human judgments. Curriculum-DPO~\cite{croitoru2025curriculum} adopts a curriculum learning approach, improving sample efficiency by gradually increasing task difficulty. PPD~\cite{dang2025personalized} extends DPO to model individual user preferences, addressing the limitations of population-level supervision.

Finally, DPO has also been applied to autoregressive models. LightGen~\cite{wu2025lightgen} uses post-training DPO to refine generation quality. Image-Generation-CoT~\cite{guo2025can} combines DPO with Chain-of-Thought (CoT) prompting, enabling preference alignment in sequential generation tasks. These efforts reflect the growing utility and flexibility of DPO for aligning image generation models with nuanced, user-centered objectives.

\subsection{GRPO-based Image Generation}
Group Relative Policy Optimization (GRPO)~\cite{shao2024deepseekmath} is a recent extension of policy optimization that offers improved generalization and training stability. As a variant of PPO, GRPO has gained traction in visual content generation and has been applied across diffusion- and flow-based pipelines.

DanceGRPO~\cite{dancegrpo} proposes a unified GRPO framework that supports diverse generative paradigms, including diffusion models and rectified flows. It is applicable across multiple tasks such as text-to-image, text-to-video, and image-to-video generation. DanceGRPO supports various backbone models (e.g., Stable Diffusion, FLUX) and reward types (e.g., aesthetics, alignment, motion), demonstrating the flexibility of GRPO in unifying heterogeneous tasks and architectures under a single policy-based training scheme.

Flow-GRPO~\cite{liu2025flow} extends GRPO to flow-based generative models by reformulating generation as a stochastic differential equation (SDE). This allows for effective RL-based exploration. It also introduces a denoising-step shrinking strategy to improve both sample efficiency and output fidelity.
Further extending its application, TALK2MOVE~\cite{tan2026talk2move} applies Flow-GRPO to instruction-based geometric image editing, utilizing a spatially-grounded reward function based on segmentation and depth estimates to achieve precise object manipulation without paired training data.

The application of GRPO also extends to autoregressive generation paradigms. T2I-R1~\cite{jiang2025t2i} introduces BiCoT-GRPO, a joint optimization framework that aligns both semantic-level and token-level Chain-of-Thought (CoT) reasoning within a single GRPO loop. SimpleAR~\cite{wang2025simplear} demonstrates that integrating GRPO with supervised fine-tuning can significantly enhance prompt-image alignment and visual quality. Building on this, ReasonGen-R1~\cite{zhang2025reasongen} further incorporates CoT reasoning into the GRPO training process to improve logical coherence in multi-step generation.

\section{RL for Video Generation}\label{sec_vid}
Recent advances in generative modeling, particularly diffusion and autoregressive architectures, have significantly improved the quality of video synthesis from text. Despite this progress, most models are still trained using surrogate objectives such as maximum likelihood estimation (MLE) or reconstruction loss. These objectives often misalign with perceptual, semantic, or temporal criteria valued by human observers. As a result, generated videos may exhibit artifacts such as motion inconsistency, semantic drift, or physically implausible dynamics.

Reinforcement learning (RL) offers a principled framework to address these limitations. It enables direct optimization of non-differentiable or preference-aligned objectives through interaction. In video generation, RL has been used to incorporate structured feedback, enforce temporal and physical consistency, and adapt to complex user intent.

This section reviews recent developments at the intersection of RL and video generation. It highlights how RL contributes to improving controllability, realism, and alignment in generative video systems. Table~\ref{tab:rl_methods_overview} summarizes representative RL methods for video generation, illustrating the field's evolution from classical policy optimization to preference-driven techniques.




\begin{table}[t!]
\centering
\caption{Comprehensive overview of reinforcement learning methods applied to video generation, organized chronologically.}
\setlength{\tabcolsep}{2pt}
\begin{tabular}{
    >{\raggedright\arraybackslash}p{3.4cm}
    >{\raggedright\arraybackslash}p{2.8cm}
    >{\raggedright\arraybackslash}p{2.5cm}
    >{\raggedright\arraybackslash}p{2.8cm}
}
\hline
\toprule
\textbf{Method} & \textbf{RL Approach} & \textbf{Publication Date} & \textbf{Venue} \\
\midrule
AdaDiff \cite{zhang2025adadiff} & Policy Gradient & Nov 2023 & AAAI 2025 \\
RLAVE \cite{hu2023reinforcement} & Actor-Critic & Nov 2024 & MM 2023 \\
InstructVideo \cite{instructvideo} & Reward Fine-tuning & Dec 2023 & CVPR 2024 \\
VADER \cite{vader} & Reward Gradient & Jul 2024 & arXiv \\
SePPO \cite{sspo}& DPO & Oct 2024 & arXiv \\
VideoAgent \cite{soni2025videoagent} & Online Policy Fine-tuning & Oct 2024 & ECCV 2024 \\
RL-V2V-GAN \cite{v2vganrl} & Policy Gradient & Oct 2024 & arXiv \\
E-Motion \cite{emotion} & PPO & Oct 2024 & NeurIPS 2024\\
Free$^2$Guide~\cite{kim2024free} & Path Integral Control & Nov 2024 & arXiv \\
IDOI \cite{furuta2024improving} & RWR, DPO & Dec 2024 & arXiv \\
FLIP \cite{gao2024flip} & Actor-Critic & Dec 2024 & ICLR2025 \\
VideoDPO \cite{videodpo} & DPO & Dec 2024 & CVPR 2025 \\
OnlineVPO \cite{zhang2024onlinevpo} & DPO & Dec 2024 & arXiv \\
VisionReward \cite{visionreward} & DPO & Dec 2024 & arXiv \\
HuViDPO \cite{jiang2025huvidpo} & DPO & Feb 2025 & arXiv \\
IPO \cite{yang2025ipo} & DPO & Feb 2025 & arXiv \\
HALO \cite{wang2025harness}  & DPO & Feb 2025 & arXiv \\
AAVG \cite{zhu2025aligning} & DPO & Apr 2025 & arXiv \\
SkyReelsV2 \cite{skyreels} & DPO & Apr 2025 & arXiv \\
Phys-AR \cite{physar} & GRPO & Apr 2025 & arXiv \\
DanceGRPO \cite{dancegrpo} & GRPO & May 2025 & arXiv \\
Diffusion-NPO \cite{wang2025diffusion}  & NPO & May 2025 & ICLR 2025 \\
RLVR-World \cite{rlvrworld} & GRPO & May 2025 & arXiv \\
InfLVG \cite{inflvg} & GRPO & May 2025 & arXiv \\
RDPO \cite{rdpo} & DPO & Jun 2025 & arXiv \\
DenseDPO \cite{densedpo} & DPO & Jun 2025 & arXiv \\
EchoMimicV3 \cite{echomimicv3} & DPO & Jul 2025 & arXiv \\
\hline
\end{tabular}

\label{tab:rl_methods_overview}
\end{table}
\subsection{Improving Diffusion Sampling Efficiency}
One practical application of RL in video generation is optimizing the diffusion sampling process to reduce computational cost without sacrificing output quality. AdaDiff~\cite{zhang2025adadiff} proposes an adaptive step selection method that accelerates sampling by learning to adjust denoising step sizes dynamically. The process is framed as an RL problem, where the state is the current noisy video frame and timestep, the action is the size of the next denoising step, and the reward reflects both generation speed and final output quality. A policy network is trained to select step sizes that balance efficiency and fidelity. Larger steps are chosen when coarse updates suffice, while smaller steps are used when fine details are required. This adaptive scheduling enables faster, high-quality video generation with fewer denoising iterations.


\subsection{Sequential Control: Planning and Editing}
Beyond low-level sampling, reinforcement learning (RL) enables high-level procedural control by formulating video creation as a sequential decision-making task. This includes both the planning of content and motion before generation, as well as the editing or refinement of existing videos to meet structured and goal-driven objectives.

In video planning, RL guides generation toward task-specific goals. FLIP~\cite{gao2024flip} formulates instructional video generation as a clip selection problem. It adopts an actor-critic framework, where the policy selects video segments to fulfill a textual instruction. A vision-language model provides dense feedback as the reward signal, allowing the system to learn semantically aligned and visually coherent video plans. VideoAgent~\cite{soni2025videoagent} approaches planning from an embodied AI perspective. It uses demonstration-based videos to bootstrap a visual planner, which is later refined by executing rollouts in real environments. Although RL underpins the learning loop, the method avoids explicit policy gradients and instead uses reward-guided data selection.

More than planning, RL is also applied in video editing and transformation. RL-V2V-GAN~\cite{v2vganrl} performs unsupervised video-to-video translation—for example, transforming summer scenes into winter. It incorporates policy gradients within a GAN framework, rewarding outputs with consistent temporal structure and style fidelity. This allows the model to learn coherent transformations without paired training data. E-Motion~\cite{emotion} focuses on future motion prediction using event-camera inputs. It frames the reverse process of diffusion as a MDP and applies PPO to fine-tune motion trajectories using perceptual rewards such as FVD and SSIM.

RL has also been extended to general-purpose video editing. RLAVE~\cite{hu2023reinforcement} proposes a framework where the actor selects editing operations—such as cuts, transitions, or clip selection—and the critic evaluates narrative coherence, pacing, and aesthetics using a vision-language model. The system is rewarded for producing stylistically consistent and engaging video edits, bridging the gap between automatic synthesis and human-guided post-production.

\begin{table}[t!]
\centering
\caption{Commonly Used Reward Models in Visual Content Generation}
\setlength{\tabcolsep}{2pt}
\begin{tabular}{
    >{\raggedright\arraybackslash}p{4.5cm}
    >{\raggedright\arraybackslash}p{2.8cm}
    >{\raggedright\arraybackslash}p{2.3cm}
    >{\raggedright\arraybackslash}p{5.1cm}
}
\hline
\toprule
\textbf{Model Name} & \textbf{Modality} & \textbf{Backbone} & \textbf{Typical Use} \\
\midrule
LAION Aesthetic Predictor~\cite{laion_aesthetic} & Image & ViT & Aesthetic quality reward \\
CLIPScore~\cite{radford2021learning} & Text-Image & ViT & Cross-modal similarity reward \\
BLIPScore~\cite{li2022blip} & Vision-Language & ViT & Semantic consistency reward \\
VideoCLIP-XL~\cite{wang2024videoclip} & Text-Video & ViT & Text-video alignment reward \\
PickScore~\cite{kirstain2023pick} & Text-Image & ViT & Pairwise preference reward \\
HPSv2~\cite{wu2023human} & Text-Image & ViT & Human preference reward \\
UnifiedReward~\cite{wang2025unified} & Image + Video & LLM & Unified generation reward \\
IPO~\cite{yang2025ipo} & Text-Video & LLM & Iterative T2V alignment reward \\
LiFT~\cite{wang2024lift} & Text-Video & LLM & Human-feedback alignment reward \\
VideoScore~\cite{he2024videoscore} & Video & LLM & Video quality assessment reward \\
VisionReward~\cite{visionreward} & Image + Video & LLM & Fine-grained VQA-based reward \\
VideoAlign\cite{liu2025improving} & Text-Video & LLM & Multidimensional T2V quality reward \\
AnimeReward~\cite{zhu2025aligning} & Image-to-Video & LLM & Anime video preference reward \\
\hline
\end{tabular}
\label{tab:reward_models}
\end{table}

\subsection{Alignment with Human Preferences}
One of the most impactful applications of reinforcement learning (RL) in video generation is post-training alignment. This stage fine-tunes pretrained generators to better match subjective human preferences. Alignment methods vary in how they incorporate reward signals—ranging from explicit policy optimization to gradient-based preference modeling. The design and quality of these reward signals are therefore fundamental. Table~\ref{tab:reward_models} provides an overview of representative reward models commonly used to guide such alignment processes.

\noindent \textbf{Policy Optimization:}
A core group of methods uses formal policy optimization to align generative models. While early RL algorithms struggled with stability, recent approaches offer more robust performance. Group Relative Policy Optimization (GRPO) is one such method. DanceGRPO~\cite{dancegrpo} applies GRPO to a variety of visual generation tasks, including text-to-image, text-to-video, and image-to-video synthesis. It reformulates both diffusion sampling and rectified flows as stochastic differential equations (SDEs), allowing GRPO to operate across architectures and training paradigms. It supports multiple foundation models (e.g., Stable Diffusion, HunyuanVideo) and various reward types, such as aesthetic quality, alignment, motion consistency, and binary feedback. This flexibility enables GRPO to effectively instill structured knowledge during post-training.





\noindent \textbf{Direct Preference Optimization:}
Direct Preference Optimization (DPO) has become the dominant framework for aligning video generation with human preferences. Compared to Reinforcement Learning with Human Feedback (RLHF), DPO is more stable and efficient. It bypasses exploration and avoids explicit reward modeling by operating on static datasets of ranked sample pairs. A simple classification loss encourages the model to replicate preferred outputs.

VideoDPO~\cite{videodpo} is an early implementation that builds large-scale preference datasets using automated scoring heuristics. HuViDPO~\cite{jiang2025huvidpo} extends this framework to text-to-video generation. It proposes a structured loss and architectural refinements—such as First-Frame Conditioning and Sparse Causal Attention—to improve visual quality and temporal consistency.

To support fine-grained alignment, several methods decompose preferences into interpretable dimensions. VisionReward~\cite{visionreward} uses Multi-Objective Preference Optimization (MPO) to balance temporal consistency, text fidelity, and realism. HALO~\cite{wang2025harness} further introduces Granular-DPO, which applies patch-level feedback to correct localized artifacts. DenseDPO~\cite{densedpo} improves structural alignment by denoising corrupted videos and using segment-level annotations for supervision. This reduces annotation cost while preserving quality.

Advances in preference data sourcing have also driven DPO’s scalability. RDPO~\cite{rdpo} uses physics-based heuristics to automatically generate preference pairs from real videos. For example, preferring a ball that falls over one that floats. This approach encodes physical plausibility without human labeling.

Some studies improve the learning dynamics of DPO itself. OnlineVPO~\cite{zhang2024onlinevpo} proposes an online variant that continuously samples, ranks, and updates the model in real time. This reduces temporal artifacts and improves subject consistency without relying on static datasets.

Moreover, DPO has also been integrated into hybrid training pipelines. SePPO~\cite{sspo} and EchoMimicV3~\cite{echomimicv3} alternate between supervised learning and preference optimization, combining the stability of supervised training with the alignment precision of DPO. VPO~\cite{vpo} extends this idea by optimizing user prompts, rather than generation models, using DPO. SkyReelsV2~\cite{skyreels} implements a multi-stage pipeline involving supervised finetuning, motion-specific DPO, infinite-length generation, and final polishing. This structured framework enables open-source cinematic video synthesis.

New formulations of the DPO objective have also emerged. AAVG~\cite{zhu2025aligning} introduces Gap-Aware Preference Optimization (GAPO), which weights training samples based on preference confidence. IDOI~\cite{furuta2024improving} combines DPO and Reward-Weighted Regression (RWR) in a unified probabilistic framework, using binary vision-language feedback to optimize dynamic object interactions. IPO~\cite{yang2025ipo} proposes Iterative Preference Optimization, an RL-based framework that alternates between evaluation and critic-guided updates. This improves semantic fidelity, motion realism, and aesthetic appeal. In contrast, Diffusion-NPO~\cite{wang2025diffusion} focuses on undesirable outputs. It introduces Negative Preference Optimization (NPO), which trains models to suppress unwanted behaviors using inverted preference labels. The negatively aligned model is used as the unconditional branch in classifier-free guidance, enhancing separation between high- and low-quality generations. This strategy improves visual quality, robustness, and human alignment across tasks.

\noindent \textbf{Reward Finetuning.}
Although distinct from classical reinforcement learning, reward finetuning encompasses a family of methods that directly supervise generative models using reward signals. These approaches typically treat the reward as an optimization objective, bypassing policy learning and value estimation. While not reinforcement learning in the strict sense, they share a common goal: aligning generated outputs with high-level human preferences.

InstructVideo~\cite{instructvideo} represents an early instance of this paradigm. It formulates video generation as an editing task and leverages pretrained image-level reward models to evaluate outputs. This lightweight strategy reduces computational overhead by iteratively applying reward feedback without full model retraining.
VADER~\cite{vader} expands this framework by integrating multiple expert models—such as CLIP for semantic alignment and YOLO for object detection—into a unified, differentiable reward function. This enables gradient-based optimization across multiple quality dimensions, improving output fidelity and relevance.

In contrast to these gradient-based methods, Free$^2$Guide~\cite{kim2024free} adopts a gradient-free approach inspired by optimal control. It uses black-box reward signals—such as evaluations from vision-language models—to guide inference-time decisions. This plug-and-play strategy conditions generation on reward signals without modifying the generator’s parameters.

\subsection{Physical Consistency and World Modeling}
A growing line of research explores how reinforcement learning can enhance physical realism and world modeling in video generation. These methods aim to produce content that not only looks plausible but also respects underlying physical principles and environment dynamics.

Phys-AR~\cite{physar} is a pioneering effort in this direction. It integrates symbolic reasoning with reinforcement learning in a post-training setting. The model first converts video frames into symbolic tokens using a Diffusion Timestep Tokenizer (DDT). It then frames generation as a token-level MDP and applies GRPO to refine the generation policy. A key contribution of Phys-AR lies in its reward design, which encodes physical laws such as velocity consistency and mass-informed motion. This encourages the model to generate trajectories that exhibit uniform motion, parabolic arcs, and realistic collision responses.

RLVR-World~\cite{rlvrworld} offers a complementary approach, addressing a fundamental limitation of MLE in video generation. MLE often produces outputs that are blurry or semantically unstable, as it does not optimize for perceptual quality. RLVR-World introduces a reinforcement learning-based fine-tuning scheme that directly targets perceptual rewards, including LPIPS, SSIM, and task accuracy. GRPO is again employed to align the generation policy with these metrics. Importantly, this framework extends beyond video, demonstrating applicability to language modeling and other generative tasks.

\subsection{Inference-Time Reinforcement Learning}
While RL is typically applied during training for alignment or control, recent work has explored its use at inference time to dynamically guide generation decisions. This is especially useful for long-form video synthesis, where models must maintain temporal consistency, semantic coherence, and prompt fidelity over extended sequences.

InfLVG~\cite{inflvg} demonstrates this paradigm by incorporating GRPO directly into the inference loop. Instead of relying on static context selection, InfLVG formulates generation as a sequential decision process. At each step, the model samples a set of candidate continuations and scores them using a composite reward function. This reward balances three key aspects: (1) content consistency, measured by face identity similarity; (2) prompt relevance, assessed via CLIP-based text–video alignment; and (3) artifact suppression, penalized using a vision–language classifier.
These scores are used to compute relative advantages among candidates. GRPO then updates the policy via a clipped objective, guiding a Plackett–Luce top-K sampling procedure to select the most appropriate context tokens. This inference-time optimization enables adaptive context selection while keeping the computational budget fixed. As a result, InfLVG achieves significant length extension—up to 9× longer—while improving visual sharpness and temporal consistency.

In summary, we provide a comparative analysis of representative approaches. Owing to the heterogeneity of experimental settings and the variability of reported evaluation protocols, we restrict our discussion to methods that have been quantitatively assessed on the VBench benchmark. The corresponding evaluation results are reported in Table~\ref{tab:metrics}.

\begin{table}[t]
\centering
\caption{Evaluation on selected VBench sub-metrics.}
\setlength{\tabcolsep}{2pt}
\begin{tabular}{lcccccc}
\toprule
\multirow{2}{*}{\textbf{Models}} & \textbf{Image} & \textbf{Multiple} & \textbf{Human} & \textbf{Motion} & \textbf{Subject} & \textbf{Aesthetic} \\
 & \textbf{Quality} & \textbf{Objects} & \textbf{Action} & \textbf{Smoothness} & \textbf{Consistency} & \textbf{Quality} \\
\midrule
VC2+VideoDPO \cite{videodpo} & - & 52.29 & 99.00 & 92.18 & 95.69 & 63.18 \\
Turbo+VideoDPO \cite{videodpo} & - & 51.98 & 94.00 & 88.85 & 96.10 & 68.98 \\
CogVid+VideoDPO \cite{videodpo} & - & 54.04 & 81.00 & 88.64 & 94.67 & 58.64 \\
VisionReward \cite{visionreward} & - & 71.54 & 98.40 & - & - & - \\
Turbo-v1+HALO \cite{wang2025harness} & 72.07 & 54.97 & 95.00 & - & - & - \\
Turbo-v2+HALO \cite{wang2025harness} & 69.11 & 67.5 & 97.6 & - & - & - \\
CogVidX-2B+HALO \cite{wang2025harness} & 61.90 & 72.91 & 98.00 & - & - & - \\
RDPO \cite{rdpo} & 65.11 & - & - & 99.27 & 97.04 & 55.01 \\
CogVidX-2B+VPO \cite{vpo} & - & 70.17 & 99.00 & - & - & - \\
CogVidX-5B+VPO \cite{vpo} & - & 75.73 & 99.60 & - & - & - \\
AAVG \cite{zhu2025aligning} & 68.98 & - & - & 99.13 & 95.20 & 54.28 \\
OnlineVPO \cite{zhang2024onlinevpo} & 67.36 & - & - & 99.36 & 97.58 & 55.37 \\
InstructVideo \cite{instructvideo} & 70.09 & - & - & 96.76 & 96.45 & 50.01 \\
VADER \cite{vader} & 66.08 & - & - & 98.89 & 95.53 & 53.43 \\
CogVidX-2B+IPO \cite{yang2025ipo} & 62.87 & - & - & 98.17 & 96.79 & 62.31 \\
\bottomrule
\end{tabular}
\label{tab:metrics}
\end{table}

\section{RL for 3D Content Generation}\label{sec_3d}
\begin{figure}[t!]
    \centering
    \includegraphics[width=0.45\textwidth]{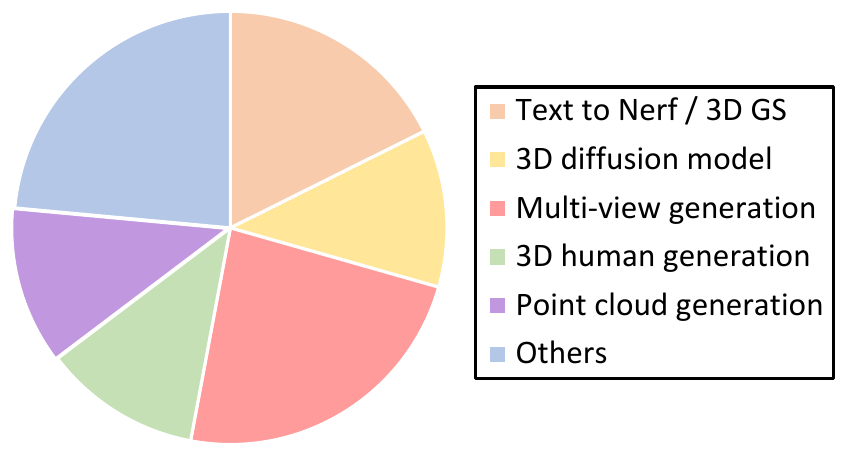}
    \caption{This figure presents the proportional breakdown of recent research topics applying RL to 3D generation, including Text-to-NeRF/3D Gaussian Splatting, 3D diffusion models, multi-view consistency optimization, human motion synthesis, point cloud modeling, and others. The balanced distribution across these areas reflects a rapidly emerging and diversifying field, where no single paradigm dominates. This suggests that RL is being broadly explored as a general-purpose optimization tool across the full spectrum of 3D generation tasks.}
    \label{fig:pie_chart}
\end{figure}

The integration of reinforcement learning (RL) into 3D content generation has introduced a powerful paradigm that blends sequential decision-making with geometric modeling. This synergy enables more adaptive, controllable, and preference-aligned synthesis of complex 3D structures. As illustrated in Figure~\ref{fig:pie_chart}, recent research efforts in this domain are broadly distributed across multiple subfields, including Text-to-NeRF/3D Gaussian Splatting, 3D diffusion modeling, multi-view consistency optimization, human motion generation, point cloud reconstruction, and other emerging directions. The balanced distribution among these topics underscores the breadth and growing maturity of RL applications in 3D generation. Unlike earlier stages dominated by a single task or representation, current trends suggest that RL is being explored as a general-purpose optimization framework across the entire 3D content generation pipeline. An overview of recent representative studies is summarized in Table~\ref{tab:3d_table}.

To organize this growing body of work, we categorize recent advances into three core areas: (1) general-purpose 3D content generation, (2) multi-view model enhancement, and (3) domain-specific applications, such as surface completion, stereo scene reconstruction, and dynamic human motion synthesis.

\subsection{3D Content Generation}

\noindent\textbf{Point Cloud-Based 3D Generation:}
Akizuki et al.~\cite{akizuki2020generative} were among the first to explore RL for 3D generation. Their method voxelizes 3D models to define spatial boundaries. The action space corresponds to voxel extension directions, while the reward encourages topological validity and penalizes self-intersections. This setup enables the agent to generate diverse, structurally sound 3D objects by learning motion trajectories within voxel grids.

Recent works extend this idea to mesh generation. DeepMesh~\cite{zhao2025deepmesh} employs Direct Preference Optimization (DPO) in 3D autoregressive models to align generation with human preferences. However, this method depends on costly manual annotation of preference pairs. Mesh-RFT~\cite{liu2025mesh} proposes a topology-aware scoring scheme using automated metrics—Boundary Edge Ratio (BER) and Topology Score (TS)—to assess mesh quality at both object and face levels. It further introduces M-DPO and quality-aware masking to refine flawed regions and capture local geometric variations.

\noindent\textbf{Depth Image-Based 3D Generation:}
Zhang et al.\cite{zhang2023point} propose a deep RL method based on A3C to reconstruct complete 3D scenes from RGB-D inputs, even under heavy occlusion. The model selects optimal viewpoints via a learned policy and applies volume-guided inpainting for progressive semantic completion. Similarly, Lin et al.\cite{lin2020modeling} use depth images as structural references and train two RL agents—one for modifying primitives, the other for adjusting vertices. Their reward is based on the change in intersection-over-union (IoU), enabling precise geometric editing.


\noindent\textbf{Text- and Image-Conditioned 3D Generation:}
Recent advances in 3D generation have increasingly explored the use of RL to enhance controllability, geometric consistency, and human preference alignment. In particular, RL provides a natural framework for optimizing non-differentiable objectives, enabling fine-grained reward-driven adaptation of 3D assets generated from text or image prompts.

DreamReward~\cite{ye2024dreamreward} represents RL-based frameworks and constructs a large-scale preference-labeled dataset of 25,304 prompt–3D asset pairs with human feedback scores. It trains a reward model, Reward3D, to evaluate generation quality. The sampling process is formulated as an RL task, where actions correspond to updates of the 3D representation, and the reward is derived from Reward3D. By integrating this learned reward with Score Distillation Sampling (SDS)~\cite{poole2022dreamfusion}, the method improves both alignment and fidelity of 3D outputs.

However, DreamReward's reliance on a dedicated reward model introduces scalability limitations. DreamDPO~\cite{zhou2025dreamdpo} addresses this by applying Direct Preference Optimization (DPO) with large pretrained vision-language models acting as zero-shot reward functions. During training, image renderings of 3D assets are ranked using textual instructions, and these rankings guide preference-based updates to the 3D representation. This framework removes the need for manual reward supervision while leveraging the generalization ability of large multimodal models. By framing the process as offline preference optimization, DreamDPO aligns 3D generation with RL principles while avoiding the instability of dynamic exploration.

To mitigate the need for preference-labeled data, DreamCS~\cite{zou2025dreamcs} introduces a scalable alternative using unpaired 3D meshes and a Cauchy–Schwarz (CS) divergence-based reward objective. It trains a geometry-aware reward model, RewardCS, on this unpaired dataset, and applies reward-guided sampling to iteratively improve the 3D output. Although not explicitly framed as an RL agent-environment loop, DreamCS still adheres to the RL paradigm by defining a reward-driven optimization landscape with structured feedback.

Recent methods have also integrated RL into the training of large-scale 3D diffusion models. GPLD3D~\cite{dong2024gpld3d} formulates two domain-specific reward functions—geometric feasibility and physical stability—and incorporates them into the generation loop to guide high-quality asset synthesis. In a more generalizable formulation, Nabla-R2D3~\cite{liu2025nabla} proposes a novel RL alignment framework that bridges 2D and 3D reward spaces. It uses a pretrained 2D reward model to guide 3D generation via a probabilistic refinement mechanism, Nabla-GFlowNet, which transforms 2D signals into structured 3D rewards. This approach highlights a key strength of RL in 3D generation: the ability to optimize across modalities using proxy signals, even in the absence of explicit 3D supervision.

\subsection{Multi-View Model Enhancement}

Beyond direct 3D asset generation, RL has been increasingly applied to improve multi-view consistency—a critical challenge in 3D generation pipelines, particularly due to the Janus problem, where different views of a synthesized object display conflicting geometry or appearance.

Carve3D~\cite{xie2024carve3d} introduces RL Fine-Tuning (RLFT) into multi-view diffusion models. Starting from a pretrained text-to-image model, Carve3D applies RLFT using a novel Multi-view Reconstruction Consistency (MRC) metric. This metric compares images generated from multiple viewpoints with views rendered from a NeRF reconstructed by a large sparse-view model (LRM). The MRC score serves as the reward signal, guiding the diffusion model to produce outputs that are both visually plausible and geometrically consistent across views.

HFDream~\cite{choi2024hfdream} advances this direction by constructing a large-scale dataset of multi-view image–text pairs annotated with human feedback. A reward model is trained to assess the viewpoint accuracy of generated images relative to textual prompts. This model is then used to fine-tune a diffusion generator via reward-weighted loss, directly aligning multi-view outputs with both human preferences and geometric fidelity in text-to-3D scenarios.

Pushing RL integration further into the image-to-3D domain, MVReward~\cite{wang2025mvreward} collects a dataset of multi-view RGB images, normal maps, and associated human preference annotations. It introduces a multi-view preference learning strategy that combines reward loss—derived from a learned preference model—with standard pretraining objectives. This hybrid RL formulation improves the structural realism and viewpoint alignment of reconstructed 3D shapes.

To alleviate the tension between high computational cost and generation quality in few-step text-to-multi-view models, Zhang et al.~\cite{zhang2025refining} propose a novel RL fine-tuning framework that reformulates the task as a constrained single-view policy optimization problem. Their method introduces ZMV-Sampling, which enhances both cross-view and text-image alignment, and MV-ZigAL, a reward-advantage learning scheme. MV-ZigAL uses advantage-weighted policy updates to internalize reward-driven improvements, enabling the base sampling policy to generate coherent multi-view images with fewer inference steps.

\subsection{Other Domain-Specific 3D Applications}
RL has not only been adopted to improve the quality and alignment of general 3D assets, but also extended to domain-specific generation tasks that require structured control, physical reasoning, or multi-object coordination. These applications further exemplify the versatility of RL as a general decision-making substrate for visual content generation.

QINet~\cite{zhang2022qinet} applies RL to 3D point cloud completion using an actor-critic framework. The agent interacts with a latent-space GAN, refining noise vectors to reconstruct complete shapes from corrupted inputs. Rewards combine geometric accuracy (IoU) and latent consistency, enabling robust completion even with large missing regions.
RLSS~\cite{ostonov2022rlss} addresses sequential 3D scene generation. It formulates scene construction as an RL problem where states encode 2D layouts and object placements, and actions select and position objects. Using PPO, the agent is trained with rewards that enforce physical constraints and design objectives. The method enables diverse and realistic indoor layouts with minimal supervision.
Zhao et al.~\cite{zhao2023synthesizing} propose a scene-aware RL framework for generating human motion in 3D environments. PPO is used to train policies over a latent motion space. States capture geometry, body pose, and interaction goals, while rewards encourage goal completion, contact accuracy, and collision avoidance. The approach supports complex interactions and fine-grained path control.
Bailando~\cite{siyao2022bailando} introduces a transformer-based RL framework for 3D dance generation. A choreographic memory encodes quantized poses, and a GPT-style actor-critic predicts future motion. Beat-alignment rewards ensure synchronization with music and consistency between body parts, producing realistic and stylistically coherent dances.
Text2Stereo~\cite{garg2025text2stereo} tackles text-to-stereo image generation. It fine-tunes Stable Diffusion on vertically stacked stereo pairs and uses RL via AlignProp. Differentiable rewards guide optimization toward stereo consistency and prompt alignment, with human preferences integrated into the feedback signal.

These examples highlight the versatility of RL in enhancing 3D generation across diverse domains. By enabling dynamic feedback and structured control, RL extends generative models to more complex and interactive 3D scenarios.

{\tiny
\begin{landscape}
\begin{longtable}[c]{p{0.1\linewidth}p{0.1\linewidth}p{0.1\linewidth}p{0.1\linewidth}p{0.1\linewidth}p{0.1\linewidth}p{0.1\linewidth}p{0.1\linewidth}}
\caption{Comprehensive overview of RL methods applied to 3D generation.}
\label{tab:3d_table}\\
\hline
\textbf{Approaches} & \textbf{Year} & \textbf{Task} & \textbf{Training Techniques} & \textbf{Actions} & \textbf{Reward} & \textbf{States} & \textbf{Performance} \\ \hline
\endfirsthead
\endhead
\hline
\endfoot
\endlastfoot
Object Generation with Design Constraint \cite{akizuki2020generative} & 2020 RE & Constraint-aware 3D Objects Generation & ML & Neighboring voxels (6 or 26 directions) moving or specific LEGO blocks placing, depending on the experiment & +1 (inside geometry), -0.5/-1 (self-intersection / outside) & Local voxel window (e.g., 5x5x5) around the agent & Generates topologically consistent, stylistically varied 3D objects and LEGO assemblies, topologically consistent designs meeting constraints \\
DeepMesh \cite{zhao2025deepmesh} & 2025 Arxiv & Mesh Generation & DPO & Vertex/face token prediction in hierarchical blocks & Human preference + geometric metrics (Chamfer Distance) & Local voxel window (e.g., 5×5×5) + patch connectivity. & SOTA in geometric fidelity, visual appeal, generation efficien and face count (30k+) \\
Mesh-RFT \cite{liu2025mesh} & 2025 Arxiv & Mesh Generation & Masked DPO & Face-level token prediction with quality-aware masking & BER (Boundary Edge Ratio) + TS (Topology Score) + HD (Hausdorff Distance) metrics & Tokenized mesh sequences + local quality masks. & SOTA in both geometric accuracy and visual fidelity \\
Point Cloud Scene Completion from Single RGB-D Image \cite{zhang2023point} & 2023 TPAMI & Colored Semantic Point Cloud Scene Completion & A3C & Next-best-view selection & Accuracy, hole-filling, and point cloud recovery & Updated point cloud at each iteration & SOTA in scene completion and scene segmentation accuracy \\
3D Shapes Modeling \cite{lin2020modeling} & 2020 ECCV & 3D Shape Modeling & IL + RL & Primitive manipulation and mesh editing & IoU-based and parsimony rewards & Shape reference, primitives/edge loops, step encoding & Structurally meaningful and regular meshes \\
DreamReward \cite{ye2024dreamreward} & 2024 ECCV & Text-to-3d Generation & RLHF & Optimizing 3D representations using human feedback & Reward3D (human preference scoring) & 3D representations & SOTA in high-fidelity, human-aligned 3D generation \\
DreamDPO \cite{zhou2025dreamdpo} & 2025 Arxiv & Text-to-3D Generation & DPO & Optimizing 3D representations via pairwise comparisons & Preference scores from reward/LMM models & 3D representations & SOTA in high-quality, controllable 3D generation \\
DreamCS \cite{zou2025dreamcs} & 2025 Arxiv & Text-to-3D Generation & CS divergence & Optimizing 3D representations via unpaired preference learning & RewardCS (3D geometry-aware scoring) & 3D representations & SOTA in high-fidelity, geometrically consistent 3D generation \\
Nabla-R2D3 \cite{liu2025nabla} & 2025 Arxiv & Text-to-3D Generation & Gradient-informed RL Finetune & Sampling and rendering multi-view images & 2D appearance and geometry scores. & Latent 3D shape representations & Propose a method Nabla-R2D3 that can effectively, efficiently and robustly finetune 3D-native generative models from 2D reward models with better preference alignment, better text-object alignment and fewer geometric artifacts \\
Carve3d \cite{xie2024carve3d} & 2024 CVPR & Text-to-Multiview Generation & On-policy Policy Gradient, KL regularization & Denoising steps in diffusion process & MRC (Multi-view Reconstruction Consistency) metric & Text prompt, timestep, noisy image & Superior multi-view consistency and NeTF reconstruction quality \\
HFDream \cite{choi2024hfdream} & 2024 OpenReview & Text-to-Multiview Generation & LoRA, SFT & Generating view-aligned images & Human-annotated view alignment score & Text prompt and diffusion model output & Improved 3D consistency and text alignment \\
MVReward \cite{wang2025mvreward} & 2025 AAAI & Image-to-Multiview Generation & MVP & Denoising steps in diffusion process & Human preference score (MVReward) & Image prompt and noisy multi-view images & Improved human-aligned 3D generation quality \\
Refining Few-Step Text-to-Multiview Diffusion \cite{zhang2025refining} & 2025 Arxiv & Text-to-Multiview Generation & MV-ZigAL, Constrained Optimization & Denoising steps in diffusion process & Joint-view scores including alignment quality, geometry quality, texture quality, and overall quality, and single-view fidelity scores & Multiview-aware MDP (Markov decision process) with text prompt and viewpoint embeddings & Improved fidelity and consistency while preserving few-step T2MV diffusion baseline efficiency \\
QINet \cite{zhang2022qinet} & 2022 TGRS & Point Cloud Completion & DDPG & Revised noise of z (clean latent code) & Weighted sum of reconstruction and latent rewards & Incomplete representation in latent space (IRLS) & SOTA in shape completeness, perceptual quality, shape resolution and category distinction \\
RLSS \cite{ostonov2022rlss} & 2022 WACV & 3D Indoor Scene Generation with Predefined Design Objects & PPO & Object placement and selection & Constraint-driven rewards that include multiple conditions such as successful condition, count of objects, and failure conditions & 2D scene representation with object metadata & High-quality, diverse scene generation with fast synthesis \\
Synthesizing diverse human motions in 3d indoor scenes \cite{zhao2023synthesizing} & 2023 ICCV & Human Motions Synthesis & PPO & Latent variables of CVAE-based generative motion primitive & Goal-reaching, foot-ground contact, and penetration avoidance & configurations of 3D scene geometry, virtual human body, and intended goals to interact with & Realistic and diverse human-scene interactions with low penetration, high foot-ground contact accuracy, and efficient task completion times, outperforming baselines in perceptual studies \\
Bailando \cite{siyao2022bailando} & 2022 CVPR & 3D Dance Generation & Actor-Critic Finetune & Predicting next dance pose & Beat-align reward and half-body consistency reward & A sequence of dance poses & SOTA in motion quality, diversity, and music-dance alignment \\ \hline
\end{longtable}
\end{landscape}}

\section{Mechanisms and Insights}\label{sec_insight}
RL offers more than just an alternative optimization strategy—it introduces a fundamentally different perspective on how generative models can be guided, adapted, and aligned. Across images, videos, and 3D/4D visual content, RL has proven to be a versatile mechanism for addressing challenges that conventional supervised objectives struggle to capture. Below, we summarize the core mechanisms by which RL contributes to the advancement of generative modeling, followed by key insights and implications for future research.

\subsection{Mechanisms: What RL Enables}
\noindent \textbf{Optimization of Non-Differentiable Objectives: }
Many generation goals—such as aesthetic appeal, user satisfaction, or physical plausibility—cannot be expressed as differentiable losses. RL enables direct optimization of such objectives through reward signals, human feedback, or surrogate evaluations.

\noindent \textbf{Fine-Grained Control and Controllability: }
RL naturally handles sequential and conditional decisions, making it well-suited for guiding fine-grained generation processes. This allows for dynamic control over generation steps, prompts, or structures, especially in diffusion, autoregressive, or procedural models.

\noindent \textbf{Incorporation of Structured and Temporal Feedback: }
Especially in video and 3D/4D generation, RL leverages sequential feedback to enforce temporal consistency, causal dependencies, and physically valid trajectories—offering richer supervision than static paired data.

\noindent \textbf{Human Preference Alignment: }
Through preference modeling, RL provides a principled framework for aligning generation with subjective goals and user intent. This enables models to internalize feedback that is implicit, ambiguous, or task-specific.

\subsection{Insights: Broader Implications}
\noindent \textbf{RL as a Structural Component, Not Just a Training Trick: }
RL is increasingly used not just as a post-training optimization tool, but as a core part of generative system design—supporting reward modeling, policy-driven sampling, and feedback-aware scheduling.

\noindent \textbf{The Rise of Preference-Based Paradigms: }
DPO and its extensions have redefined how RL interacts with generative models, shifting the focus from exploration-heavy training to stable, scalable preference alignment. This marks a new era of sample-efficient, human-aligned generation.

\noindent \textbf{Unifying Modeling and Interaction: }
RL bridges the gap between learning to generate and learning to interact. It offers a foundation for developing generative agents that are adaptive, embodied, and aware—essential for real-world deployment in creative, scientific, or robotic settings.

\noindent \textbf{Beyond Static Objectives: Towards Multi-Objective, Multi-Agent Generation: }
As RL continues to evolve, we anticipate a shift toward multi-objective optimization, where generative models must balance trade-offs between quality, diversity, safety, and efficiency—possibly in multi-agent or cooperative learning setups.

\subsection{Future Work}
The convergence of RL and generative modeling is reshaping how we think about generation—not as a static mapping from input to output, but as an interactive, iterative, and goal-driven process. As generative systems become more autonomous and user-facing, the ability to learn from feedback, optimize complex objectives, and adapt to diverse preferences will be indispensable. RL provides the theoretical tools and algorithmic foundations to meet these demands.

\bibliography{sn-bibliography}








\end{document}